\documentclass[10pt, conference, compsocconf]{IEEEtran}

\usepackage{times}
\usepackage{epsfig}
\usepackage{graphicx}
\usepackage{amsmath}
\usepackage{amssymb}

\usepackage{epstopdf}
\usepackage[scaled=.92]{helvet}
\usepackage{amsmath}
\usepackage{url}
\usepackage{wrapfig}

\usepackage{times}

\usepackage{url}

\usepackage{amsfonts,amsmath,amssymb}

\usepackage{graphicx}

\usepackage{parskip}

\usepackage{multirow}

\usepackage{psfrag}

\usepackage{verbatim}

\usepackage{wrapfig}

\usepackage{paralist}

\let\labelindent\relax
\usepackage{enumitem}
\setdescription{leftmargin=0.25cm}

\usepackage[usenames,dvipsnames]{color}
\usepackage[dvipsnames]{xcolor}

\newcounter{mycounter}

\usepackage[]{algorithm2e}

\makeatletter
\newcommand{\Removelatexerror}{\let\@latex@error\@gobble}
\makeatother

\newcommand{\myalgorithm}{%
\SetInd{0.5em}{0.5em}
\let\oldnl\nl
\newcommand{\nonl}{\renewcommand{\nl}{\let\nl\oldnl}}
\newcommand{\pushline}{\Indp}
\newcommand{\popline}{\Indm\dosemic}
\begingroup
\Removelatexerror 
\begin{algorithm*}[H]
\end{algorithm*}
\endgroup}

\newcommand{\ignore}[1]{}

\def\rev#1{#1}

\newcommand{\com}[1]{{}}

\newcommand{\be}{\mathbf{e}}

\newcommand{\bn}{\mathbf{n}}

\newcommand{\bq}{\mathbf{q}}

\newcommand{\bt}{\mathbf{t}}

\newcommand{\bD}{\mathbf{D}}

\newcommand{\bI}{\mathbf{I}}

\newcommand{\bR}{\mathbf{R}}
\newcommand{\bS}{\mathbf{S}}

\newcommand{\mL}{\mathcal{L}}

\makeatletter
\DeclareRobustCommand\onedot{\futurelet\@let@token\@onedot}
\def\@onedot{\ifx\@let@token.\else.\null\fi\xspace}

\def\eg{\emph{e.g}\onedot} 
\def\ie{\emph{i.e}\onedot}

\def\etal{\emph{et al}\onedot}
\makeatother

\usepackage[pagebackref = true,
            breaklinks = true,
            letterpaper = true,
            bookmarks = false,
            colorlinks = true,
            linkcolor = red,
            urlcolor  = blue,
            citecolor = green,
            anchorcolor = blue]{hyperref}

\begin{document}

\title{3D Shape Reconstruction from Sketches via Multi-view Convolutional Networks \vspace{-3mm}}
\author{
    \IEEEauthorblockN{
        Zhaoliang Lun \hspace{5mm}
        Matheus Gadelha \hspace{5mm}
        Evangelos Kalogerakis \hspace{5mm}
        Subhransu Maji \hspace{5mm}
        Rui Wang}
    \IEEEauthorblockA{University of Massachusetts Amherst}
}

\maketitle

\begin{abstract}
We propose a method for reconstructing 3D shapes from 2D sketches in the form of line drawings. Our method takes as input a single sketch, or multiple sketches, and outputs a dense point cloud representing a 3D reconstruction of the input sketch(es). The point cloud is then converted into a polygon mesh. At the heart of our method lies a deep, encoder-decoder network. The encoder converts the sketch into a compact representation encoding shape information. The decoder converts this representation into depth and normal maps capturing the underlying surface from several output viewpoints. The multi-view maps are then consolidated into a 3D point cloud by solving an optimization problem that fuses depth and normals across all viewpoints.  Based on our experiments, compared to other methods, such as volumetric networks, our architecture offers several advantages, including more faithful reconstruction, higher output surface resolution, better preservation of  topology and shape structure.
\vspace{-2mm}

\end{abstract}



\vspace{-3mm}
\section{Introduction}
\label{sec:introduction}
\vspace{-2mm}
We consider the problem of 3D shape reconstruction from sketches. Contours in a sketch convey important characteristics of the underlying shape such as its figure-ground boundaries, surface curvature, and occlusions~\cite{koenderink1984does,waltz1975understanding,malik1987interpreting}. They are also commonly used by artists in the initial stages of character design and object modeling due to the relative ease of sketching. However, the process of converting sketches to a 3D model is time consuming and cumbersome. 

We propose an architecture to infer a 3D shape that is consistent with sketches from one or more views of an object. 
Our method is based on a Convolutional Network (ConvNet) trained to map sketches to 3D shapes. Although ConvNets have been successfully applied to a number of  image modality transformation tasks~\cite{larsson2016learning,zhang2016colorful, johnson2016perceptual,ulyanov2016texture,pix2pix2016}, their use for explicit 3D shape generation poses numerous challenges. Most prior work has used voxel-based representations for 3D shapes~\cite{wu2016learning,choy20163d,yan2016perspective}. 
However, this scales poorly with the resolution of the voxel grid. 3D shapes can be instead efficiently represented through surface-based representations, such as  polygon meshes. However, it is difficult to parameterize meshes in a consistent manner such that they are generated by ConvNets, and unlike voxels, they are not amenable to convolutions over regular grids.
Thus, their applicability has been limited to categories (\eg, faces, human bodies) where
surface elements
can be consistently parameterized
through correspondence techniques and generated through simple generative models~\cite{allen2003space,blanz1999morphable,cashman2013shape,Huang:2015:deeplearningsurfaces}.

In this work we instead adopt a multi-view architecture for 3D shape reconstruction inspired by recent work showing that ConvNets have the ability to model geometric and viewpoint transformations of an object~given natural images \cite{dosovitskiy2015learning, tatarchenko2015single,tatarchenko2016multi,yang2015weakly,zhou2016view}.
However, unlike prior multi-view  synthesis works, we consider the \emph{full pipeline} of 3D shape reconstruction, and also condition it on 
line drawings, which are  more challenging inputs than natural images due to the lack of shading or color information. Our approach is based on minimizing a joint energy function over input sketches, multi-view depth and surface normals, and point clouds. Our inference algorithm obtains a   set of \emph{depth maps} and \emph{surface normals} of the shape from a collection of viewpoints using a \emph{feed-forward network}. We then infer a dense \emph{point cloud} that is consistent with the predicted depths and normals across all the viewpoints by minimizing our energy function. The point cloud is then converted to a discretized surface in the form of a polygon mesh and optionally further optimized to match the input line drawings more precisely.

Our approach appears to be the first that considers a  learned, view-based representation for \emph{generating} 3D shapes from sketches. The view-based representation allows us to process depth and normals at a considerably higher resolution and speed compared to voxel-based representations on existing hardware. Moreover, by incorporating the best of feed-forward architectures and mesh-based representations we are able to predict 3D shapes at a significantly higher quality. Finally,  our architecture is trained on automatically generated, synthetic sketches of 3D shapes without requiring  supervision in the form of human line drawings.
Once trained, our method  can generalize to reconstruct 3D shapes from   human line drawings that can be approximate, noisy and not perfectly consistent across different viewing angles. Finally, as a by-product of our training procedure, our network also provides   descriptors that can be used to perform sketch-based shape retrieval from 3D\ model collections.
On two qualitatively different datasets (character models and man-made objects), our proposed approach achieves significantly better reconstruction results than alternative approaches in terms of several metrics (Hausdorff distance, Chamfer distance, voxel intersection over union,  errors in depth and normal maps) and also based on a  user study.

\vspace{-2mm}
\section{Related Work}
\label{sec:related_work}
\vspace{-2mm}
\paragraph{3D geometric inference from line drawings} 
Compared to using natural images, estimating 3D shape from line drawings is considerably more challenging due to the lack of shading or texture information. 
Early works~\cite{waltz1975understanding,malik1987interpreting,Lipson96,Zeleznik:1996:SIS} formulate the process of inferring a 3D shape based on reasoning about local geometric properties, such as convexity, parallelism, orthogonality and discontinuity, implied by lines and their intersections (``junctions"), to find a globally consistent shape. 
These approaches produce reasonable geometry when applied to specific families of polyhedral objects, but are less effective for organic shapes with smoothly varying surfaces. For smooth shapes, hand-designed rules  are usually devised to extrude or elevate a 3D surface from contours \cite{Igarashi:1999:TSI,Olsen}. More recent methods enable the creation of freeform surfaces by exploiting geometric constraints present in specific types of line drawings, such as polyhedral scaffolds, cross-section lines and curvature flow lines \cite{Schmidt:2009:ADS,Xu:2014:True2Form,Pan:2015:FAS}. All these methods derive geometric constraints from specific types of lines,  require very accurate input drawings, and  can only reconstruct what is drawn. On the other hand, various studies~\cite{koenderink1992surface,cole2009well} showed that humans can consistently interpret  3D shapes  from sparse and approximate line drawings (up to a \emph{bas-relief} transformation~\cite{belhumeur1999bas}). Although the exact mechanism of 3D shape perception in humans is not well understood, this indicates that pure geometric-based methods may not be able to  mimic the human ability of\ shape understanding from sketches.
\vspace{-2mm}
\paragraph{Learning-based methods for shape synthesis}
In contrast to pure geometric methods, learning-based approaches argue that shape interpretation is  fundamentally a learning problem, otherwise it is highly under-constrained. A large number of learning-based methods have focused on estimating 3D shapes from single, natural images that include color and texture. Early work was based on analyzing shading and texture cues within image regions~\cite{hoiem2005geometric,saxena2009make3d}, while more recent work has employed ConvNets for predicting surface depth and normals from real images~\cite{eigen2015predicting,wang2015designing}. Driven by the success of  \emph{encoder-decoder} architectures~\cite{larsson2016learning,zhang2016colorful, johnson2016perceptual,ulyanov2016texture,pix2pix2016}
that can effectively map inputs from one domain to another, newer methods use such architectures  with convolutions in three dimensions to generate 3D shapes in a voxelized representation~\cite{wu2016learning,choy20163d,yan2016perspective,Hane17,Riegler17,Tatarchenko17}. A different line of work has employed ConvNets to model geometric transformations of an object to predict novel viewpoints~\cite{dosovitskiy2015learning,tatarchenko2015single,yang2015weakly,zhou2016view}. The approach of Tatarchenko \etal~\cite{tatarchenko2016multi} is most related to ours. Their approach takes as input a single natural image and a viewpoint  and uses a ConvNet to predict the color and depth from the provided viewpoint. They show compelling 3D reconstructions for chairs and cars from a single color image by projecting the depth maps from multiple views into a 3D space. Our approach is inspired by this work, but differs in a number of ways. Our method operates on line drawings, a  more challenging type of input due to the lack of shading or color information. It predicts both normals and depth across multiple viewpoints, which are then integrated into a high-quality surface mesh representation 
through a joint optimization procedure. It also adapts a U-net architecture \cite{pix2pix2016} along with multi-view decoder branches and a structured loss function  to resolve ambiguities in the input line drawing. Finally, we provide a detailed comparison of view-based and voxel-based reconstruction approaches in terms of 3D shape evaluation metrics and a perceptual user study on various categories.
\vspace{-2mm}
\paragraph{Sketch-based 3D shape retrieval}
Sketch-based retrieval methods typically transform features of the input sketch and 3D shapes into a common space where comparisons can be made. Early work was  based on hand-engineered descriptors  \cite{Funkhouser:2003:SEM,pu20052d,hou2006sketch,Lee:2008:SSC,eitz2010sketch,Xu13sig,Xie13,Schneider:2014:SCC,GuoSGP2016}, while more recently, ConvNets have been proposed to learn powerful representations for sketch-based retrieval \cite{su2015multi,wang2015sketch}. Unfortunately, these methods only allow retrieval of existing  3D  shapes or parts. They provide no means to synthesize novel shapes or parts from scratch. A few recent approaches employ category-specific, predefined \emph{parametric models} to guide shape reconstruction through ConvNets \cite{Nishida:2016:ISU,huang2017shape,HanGY17}. These methods are only able to recover specific shape parameters or rules from input sketches.  If a drawing  depicts a shape that cannot be described by the parameters of these models, then the reconstruction fails. In contrast, our method learns a representation capable of predicting shapes from sketches without any predefined parametric model. We expect 3D shape priors to automatically emerge in our deep network.

\vspace{-5mm}
\section{Method}
\label{sec:method}
\begin{figure*}[t]
\centering
\includegraphics[width=0.97\linewidth]{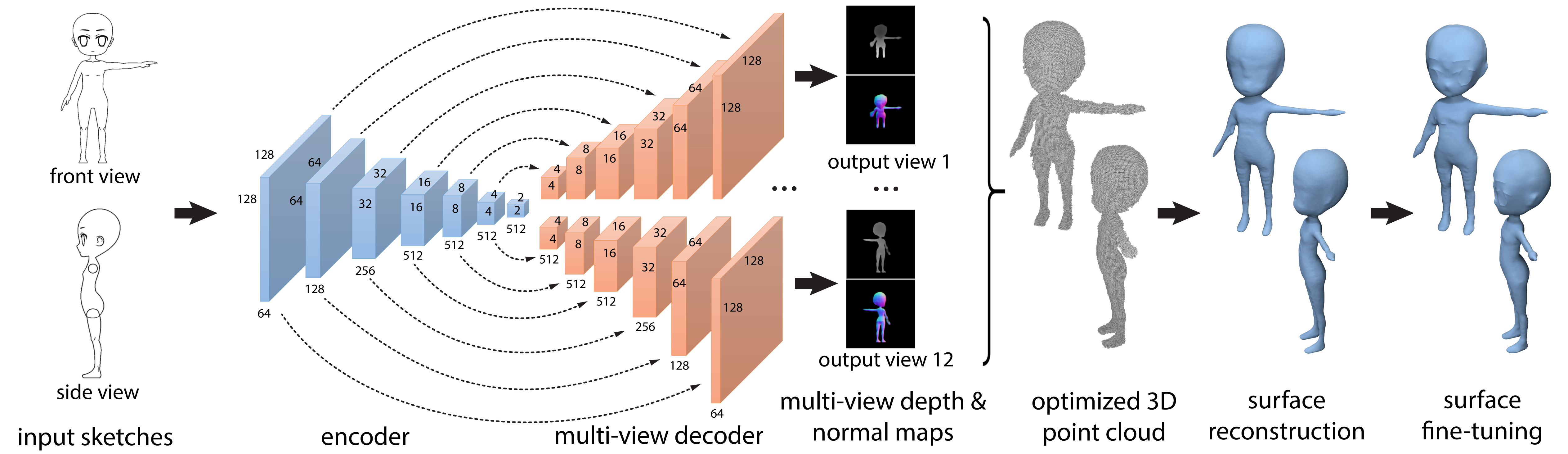}
\vspace{-2mm}
\caption{Our method takes  line drawings  as input and converts them into multi-view  surface depth and normals maps from several output viewpoints via an encoder-multi-view-decoder architecture. The maps are fused into a coherent 3D point cloud, which is then converted into a surface mesh. Finally, the mesh can be further fine-tuned to match the input drawings more precisely through geometric  deformations. }
\vspace{-5mm}
\label{fig:architecture}
\end{figure*}
\vspace{-5mm}
Given a single, or multiple, hand-drawn sketches in the form of line drawings, our method aims to reconstruct a 3D shape. Line drawings are made by humans to convey shape information \cite{decarlo2003suggestive,cole2009well}. They typically contain  external contours (silhouettes) and internal contours to underlie salient shape features. We designed a deep network to automatically translate line drawings into 2D\ images representing surface depth and normals across several output viewpoints (Figure \ref{fig:architecture}). The depth and normal predictions are  then fused into a  3D point cloud, which is in turn converted into a polygon mesh. Although surface normals could be inferred
by depth alone, we found that best reconstructions are achieved when both depth and normal predictions are made by the network and coherently fused into the  point cloud.

Our network is trained to reconstruct multi-view depth and normal maps from either a single sketch depicting the shape from a particular input view (\eg, front, side, or top), or from multiple sketches depicting the shape from different views (\eg, front and side). A single sketch may not be sufficient to reconstruct the shape accurately, \eg, the front side of an airplane does not explicitly convey information about its back.
Hence, we consider the case where users  provide multiple sketches as input at once, or provide them progressively while being guided by the intermediate shape reconstructions. 
In the latter case,
users draw from one view, then our network, which is trained to reconstruct from that view, yields a 3D shape. Users can then draw a second sketch from another view, on top of the generated shape rendered semi-transparently from that view, similar to ShadowDraw \cite{Lee:2011:SRU} (see also our supplementary material for an example). Given the previous and new sketches as input, our network, trained to reconstruct from both views, yields an updated 3D shape. The process continues until users are satisfied with the result, at which point they may edit the mesh directly. In what follows, we discuss our network architecture (Section \ref{sec:architecture}) and training (Section \ref{sec:training}). Then we discuss our optimization step to fuse the multi-view depth and normal maps into a single, coherent 3D point cloud and  its conversion to a polygon mesh (Section \ref{sec:meshing}).

\vspace{-2mm}
\subsection{Network Architecture}
\label{sec:architecture}
\vspace{-3mm}
Our ConvNet takes as input line drawings from particular  views of an object and outputs depth and normal maps in several, uniformly sampled output viewpoints (Figure \ref{fig:architecture}). Our implementation uses 12 output viewpoints located at the equidistant vertices of a regular icosahedron.
A camera is placed at each icosahedron vertex looking towards the center of the object and oriented towards the upright axis.  All our training shapes are normalized such that they fit inside the icosahedron and are also consistently oriented.

\vspace{-1mm}
\paragraph{Input} The input to our network are $256 \times 256$ intensity images representing the line drawings. When $C$ input sketches are available, they are concatenated as channels resulting in $256 \times 256 \times C$ dimensional input. For each input view configuration, we train a different network \ie, given a sketch representing the front of the object, we use the network trained to reconstruct the 3D shape from the front; or given two sketches representing the front and the top of the object, we use the network trained to reconstruct from the front and top (in this case, the two sketches are concatenated in this order). At first, this might seem  restraining, yet we note that in many traditional CAD systems, it is common for users to use canonical views \cite{Rivers:2010:MS}, and that  better reconstruction results are achieved when the network is trained to reconstruct from specific rather than arbitrary views. 
\vspace{-1mm}
\paragraph{Encoder} 
The encoder network consists of a series of convolutional layers, all using kernel size of $4$ and stride of $2$. The filter size and number per layer is shown in Figure \ref{fig:architecture}. All layers use batch normalization and leaky ReLUs (slope = $0.2$) as activation functions. The output of the encoder is a $2 \times 2 \times 512$ representation, which encodes shape information based on the input sketch(es). We note that this  representation  can be  used for sketch-based shape retrieval. 

\vspace{-1mm}
\paragraph{Decoder}
The decoder consists of 12 branches, each containing a series of  upsampling  and convolutional layers.  The branches have the same layer structure but do not share parameters. Each branch takes as input the encoder's representation and outputs a $256 \times 256 \times 5$ image for a corresponding output viewpoint.  The $5$-channel image  includes a depth map ($1$ channel), a normal map ($3$ channels  constrained  to be unit norm) and a foreground probability map for that viewpoint. All pixels with probability more than $50\%$ for foreground yield a binary mask
indicating the projected surface area under that viewpoint. The output depth and normal maps are masked using this binary mask. Following the U-net architecture \cite{Ronneberger15}, the input to each convolutional layer is formed by the concatenation of the previous layer output
in the decoder, and a corresponding layer output in the encoder (see Figure \ref{fig:architecture}). The upsampling layers of the decoder upsample their input with a factor of $2$. The convolutional layers use kernel size of $4$ and stride of $1$. Each convolutional layer is followed by batch normalization and leaky ReLU (slope = $0.2$) as activation function.  The first 3 layers in each decoder branch use dropout for regularization. The number and size of filters per layer in the decoder are shown in Figure \ref{fig:architecture}. The output layer uses the tanh activation function since depths and normals lie in range $[-1,1]$. Finally, the normal maps pass through an $\ell_2$ normalization layer that ensures they are unit length.

\vspace{-3mm}
\subsection{Training}
\label{sec:learning}
\label{sec:training}
\vspace{-2mm}
To train
our network, we need a dataset that includes 3D\ shapes along with corresponding training sketches. To
create such dataset,
one option would be to ask human subjects to provide us with line drawings depicting training 3D\ shapes. However, gathering human line drawings is  labor-intensive and time-consuming. In contrast, we generated synthetic line drawings that approximate human line drawings based on well-known  principles. Below we discuss the procedure we followed for sketch generation, then we discuss the objective used for training our network.

\vspace{-1mm}
\paragraph{Generating training sketches} Non-photorealistic rendering algorithms can be used to create synthetic line drawings of 3D shapes. 
First, contours, or silhouettes, can be estimated by finding and connecting the  set  of  points on  the  surface whose normal vector is perpendicular to the viewing direction \cite{decarlo2003suggestive}. Second, suggestive contours are extensions of contours that can be used to draw internal feature curves in shapes. These are found from zero-crossings of the radial curvature (surface curvature along viewing directions) \cite{decarlo2003suggestive}. Other types of internal feature curves include ridges and valleys, which are formed by the minima or maxima of the surface principal curvature values \cite{Ohtake:2004:RLM},
or view-dependent curvature (in this case, the lines are called ``apparent'' ridges \cite{apparentridges}). Another type of line drawings can be created through edge-preserving filtering \cite{GastalOliveira2011} applied on  images of shapes  rendered under a simple shading scheme (e.g., Phong shading) \cite{Phong:1975:ICG}. All these feature curve definitions do not necessarily coincide each other  \cite{cole2008people}. 
We use a combination of these techniques to create several variants of line drawings per input shape. This also serves as a form of data augmentation. Specifically, for each shape and input view, we create 4 synthetic sketches by using: (i) silhouettes alone, (ii) silhouettes and suggestive contours, (iii)  silhouettes, suggestive contours, ridges, valleys and apparent ridges, (iv) and edge-preserving filtering on  rendered images of shapes. 
All training sketches and corresponding ground-truth depth and normal maps are rendered under orthographic projection according to our output viewpoint setting. Using perspective projection could also be an option, however, since  depth has  a relatively short range for our rendered objects, the differences in the resulting images tend to be small. 

\vspace{-1mm}
\paragraph{Loss function} Given training sketches of shapes along with the corresponding foreground, depth and normal maps for our output viewpoints, we attempt to estimate the network parameters to minimize a loss function. 
Our loss function  consists of four terms penalizing (a) 
differences between the training depth maps and predicted depth maps,  (b) angle differences between the training normal maps and predicted normal maps, 
(c) disagreement between ground-truth and predicted foreground masks, (d) large-scale structural  differences between the predicted maps and the training maps. Specifically, given $T$ training sketches along with ground-truth foreground, depth and normal maps for our $V$ output viewpoints, our loss function is a combination of the following terms described  in the following paragraphs:
\vspace{-2mm}
\begin{align}
\!\!L\!=\!\!\sum\limits_{t=1}^T\!\!\left(\lambda_1L_{depth} (t)\!+\!\!\lambda_2L_{normal}(t)\!+\!\!\lambda_3L_{mask}(t)\!+\!\!\lambda_4L_{adv}(t)\right) \nonumber
\end{align}
\vskip -3mm
where $\lambda_1=1.0,\lambda_2=1.0,\lambda_3=1.0,\lambda_4=0.01$ are weights tuned in a hold-out validation set.

\vspace{-1mm}
\paragraph{Per-pixel depth and normal loss}  The first two terms consider  per-pixel differences in the predicted depths and normals with respect to ground-truth. Specifically, we use $\ell_1$ distance for depths and angle cosine differences for normal directions. The depth and normal differences are computed only for pixels marked as foreground in the ground-truth:
\vspace{-1mm}
\begin{align}
L_{depth} (t)= \sum_{p,v} \left(| d_{p,v}(\bS_t) - \hat{d}_{p,v,t}|\right) \hat{f}_{p,v,t} \nonumber \\
L_{normal} (t)= \sum_{p,v} \left(1 - \bn_{p,v}(\bS_t) \cdot \hat{\bn}_{p,v,t}\right) \hat{f}_{p,v,t} \nonumber
\end{align}
\vskip -5mm
where $\bS_t$ is a training sketch, $\hat{d}_{p,v,t}$ and $\hat{\bn}_{p,v,t}$ are ground-truth depth and normal for the pixel $p$ in viewpoint $v$. Each pixel has a ground-truth binary label $\hat{f}_{p,v,t}$, which is $1$ for foreground, and $0$ otherwise. The depth and normal predictions for the sketch $\bS_t$ are denoted as $d_{p,v}(\bS_t)$  and $\bn_{p,v}(\bS_t)$ respectively. We note that all training depths are normalized within the range $[-1,1]$ while predicted depths are also clamped in this range. Thus both terms above have comparable scale (\ie, both range between $[0,2]$ per pixel). We also note that we tried $\ell_2$ distance for penalizing depth differences but this tended to produce less sharp maps.

\vspace{-1mm}
\paragraph{Mask loss} Penalizing disagreement between predicted and ground-truth foreground labeling can be performed via the cross-entropy function commonly used in classification.

\vspace{-1mm}
\paragraph{Adversarial loss} We also penalize structural differences in the output maps with respect to ground-truth through an ``adversarial'' network. This has been shown to serve as an effective prior for various image-to-image transformation tasks~\cite{pix2pix2016}. The adversarial loss term takes as input a 5-channel image $\bI$ that concatenates the depth channel, the 3 normal channels, and foreground map channel produced by the decoder per viewpoint, and outputs the probability for these maps to be ``real'': $\mL_{\text{adv}} = - \sum_v \log P(\text{``real"}|\bI).$
The probability is estimated using the ``adversarial'' network trained to discriminate ground-truth (``real'') maps $\hat{\bI}$ from generated (``fake'') maps  $\bI$. Both networks are trained alternatively using the technique of~\cite{Goodfellow:2014:GAN}. The adversarial network architecture is the same as the encoder except the last layer that maps the output to probabilities via a fully-connected layer followed by a sigmoid activation.

\vspace{-2mm}
\subsection{Point Cloud and Mesh Generation}
\label{sec:meshing}
\label{sec:post_processing}
\vspace{-2mm}
Given multi-view depth and normal maps produced by our network at test time, our next goal is to consolidate them into a single, coherent 3D\ point cloud. The depth and normal predictions produced by the network are not guaranteed to be  perfect or even consistent i.e., the derivatives of the predicted depth might not entirely agree with the predicted normals, or the predicted depths for common surface regions  across different viewpoints might not yield exactly the same 3D\ points. Below we discuss an optimization approach to fuse all multi-view depth and normal map predictions into a coherent 3D point cloud, then we discuss mesh generation and post-processing to match the input sketches more precisely. Our optimization approach shares similarities with bundle adjustment and multi-view reconstruction \cite{Triggs:1999:BAM,MVSreconstruction}. In our case, our output viewpoints are fixed and we use the normal maps in our energy minimization to promote consistency between depth derivatives and surface normals.   

\vspace{-1mm}
\paragraph{Multi-view depth and normal map fusion} The  first step of the fusion process is to map all foreground pixels to 3D points.  Each pixel is  considered foreground if its predicted probability in the foreground map is above $50\%$. Given the depth $d_{p,v}$ of a foreground pixel $p$ with image-space coordinates $\{p_x, p_y\}$ in the output map of a viewpoint $v$, a 3D point $\bq_{p,v}$ can be generated according to the known extrinsic camera parameters (coordinate frame rotation $\bR_{v }$ and translation $\be_v$ in object space). Under the assumed orthographic projection, the point position is computed as:
\vspace{-1mm}
\begin{equation}
\bq_{p,v} = \bR_{v} \left[ \kappa p_x \quad \kappa p_y \quad d_{p,v} \; \right]^T + \be_v \nonumber
\end{equation}
\vskip -3mm
where $\kappa$ is a known scaling factor, representing the distance between two adjacent pixel centers when their centers are mapped to object space. Each point is also equipped with a normal $\bn_{p,v}$ based on the predicted normal map. The result of this first step is a generated point set per view. In a second step, we run ICP \cite{icp} to  rigidly align all-pairs of  point sets, which helps dealing with inconsistencies in the predicted depth maps.

\begin{figure}[t!]
\centering
\includegraphics[width=\linewidth]{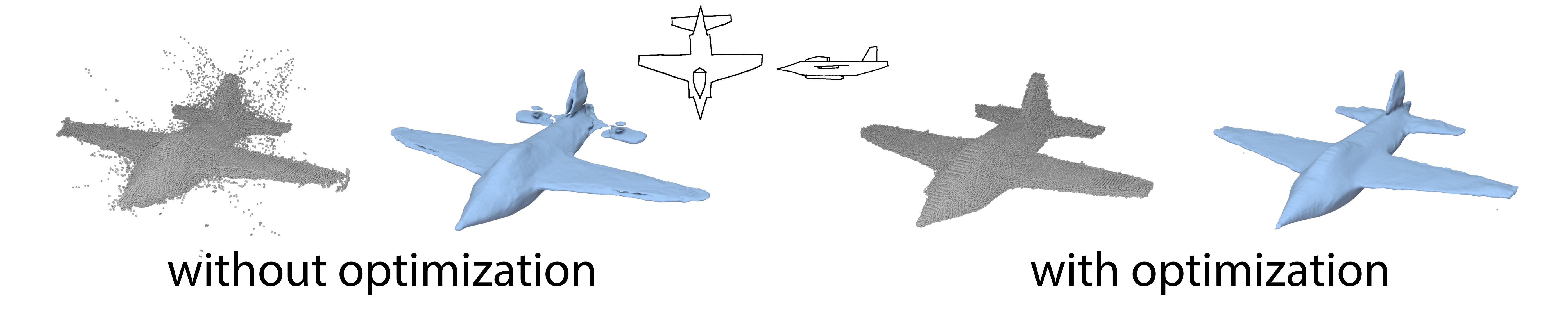}
\vspace{-7mm}
\caption{ Without optimization the noisy point cloud will lead to misaligned regions in the reconstructed shape. }
\label{fig:fusion}
\vspace{-6mm}
\end{figure}

A naive reconstruction method would be to simply concatenate all aligned point sets   from all output views into a single point cloud. However, such approach often results in a  noisy point cloud with misaligned regions due to the remaining depth map inconsistencies not handled by ICP. The effect of these inconsistencies tends to be amplified during mesh generation, since  a smooth surface cannot pass through all the misaligned regions (Figure \ref{fig:fusion}). Our optimization procedure aims to deal with this problem. 
Specifically, we treat the depths of all  pixels 
 as variables we want to optimize for. The pixel depths are optimized such that (a) they are  close to the  predicted (approximate) depths produced by the network,  (b) their first-order  derivatives  yield surface tangent vectors that are as-orthogonal-as-possible to the predicted normals, (c) they are consistent with depths and normals of  corresponding 3D\ points generated  in other viewpoints. These  requirements are expressed in a single energy over all pixel depths $\bD = \{d_{p,v}\}$ with terms imposing the above three conditions, as explained in the next paragraphs:
\vspace{-1mm}
\begin{equation}
E(\bD) =E_{net}(\bD) + E_{orth}(\bD) + E_{cons}(\bD) \nonumber
\end{equation}

\vspace{-3mm}
\paragraph{Network prediction term} The  term $E_{net}(\bD)$ penalizes deviation from the approximate depths $\tilde{d}_{p,v}(\bS_t)$ produced from the network at  each pixel $p$ and viewpoint $v$:
\vspace{-1mm}
\begin{equation}
E_{net}(\bD) = w_1 \sum\limits_{p,v} ( {d}_{p,v} - \tilde{d}_{p,v}(\bS_t) )^2 \nonumber
\end{equation}
\vskip -5mm
where $w_1$ weights this term (set  to $1.0$ through hold-out validation). We use $\ell_2$  norm here so that the energy minimization yields a linear system that can be solved efficiently. \vspace{-5mm}

\vspace{0mm}
\paragraph{Orthogonality term} The term $E_{orth}(\bD)$ penalizes deviation from orthogonality between surface tangents, approximated by first-order depth derivatives, and predicted surface normals $\tilde{\bn}_{p,v}(\bS_t)$. Given a 3D point $\bq_{p,v}$ generated for pixel $p$ and viewpoint $v$, we  estimate two surface tangent directions based on  first-order depth  derivatives \cite{Nehab:2005:ECP}:
\vspace{-2mm}
\begin{equation}
\bt_{p,v}^{(x)} = \left[ \kappa \quad 0 \quad \frac{\partial d_{p,v} }{\partial x} \right]^T,\,\,\ \bt_{p,v}^{(y)} = \left[ 0 \quad \kappa \quad \frac{\partial d_{p,v} }{\partial y} \right]^T \nonumber
\end{equation}
\vskip -3mm
The derivatives can be approximated with a horizontal and vertical gradient filter that is convolved with depths in a $3 \times 3$ neighborhood around $p$. The energy term is expressed as:
\vspace{-1mm}
\begin{equation}
E_{orth}(\bD) = w_2 \sum\limits_{p,v} [( \bt_{p,v}^{(x)} \cdot \tilde{\bn}_{p,v}(\bS_t) )^2 + ( \bt_{p,v}^{(y)} \cdot \tilde{\bn}_{p,v}(\bS_t) )^2] \nonumber
\end{equation}
\vskip -3mm
where $w_2$ is a weight (set  to $1.0$ through holdout validation).
Since the  derivatives are unreliable near the shape silhouette, we omit silhouette points  for each view from this term.

\vspace{-2mm}
\paragraph{View consistency term} Given a 3D point $\bq_{p,v}$ generated from pixel $p$ at viewpoint $v$, we can calculate its depth with respect to the image plane of another viewpoint $v'$ as well as the pixel that it is projected onto as: $p'=\Pi_{v'}(\bq_{p,v})$, where $\Pi_{v'}$ denotes orthographic projection based on the  parameters of viewpoint $v'$.
When the 3D point is not occluded and falls within the  image formed at viewpoint $v'$, the calculated depth $d_{v'}(\bq_{p,v})$ of that point should be in agreement with the depth $d_{p',v'}$ stored in the corresponding pixel $p'$ of the viewpoint $v'$. Similarly, the  normal of that point  $\bn_{v'}(\bq_{p,v})$ relative to the viewpoint $v'$ should be as-orthogonal-as possible to  the surface tangent vector, approximated by the derivative of the depth stored in the corresponding pixel  $p'$. The view consistency term penalizes: (a) squared differences between the depth at each pixel and the calculated depth of all 3D points projected onto that pixel, (b) deviation from orthogonality between the surface tangent vector at each pixel and the normal of all 3D points projected onto that pixel. The term is expressed as follows:
\vspace{-1mm}
\begin{align}
&\ E_{cons}(\bD) = w_3 \sum\limits_{\substack{p,v,p',v':\\p'=\Pi_{v'}(\bq_{p,v})}}  ( {d}_{p',v'} - d_{v'}(\bq_{p,v}) )^2 + \nonumber \\
& + w_4 \sum\limits_{\substack{p,v,p',v':\\p'=\Pi_{v'}(\bq_{p,v})}} 
\!\!\!\!\!\! ( \bt_{p',v'}^{(x)} \cdot \bn_{v'}(\bq_{p,v}))^2 + ( \bt_{p',v'}^{(y)} \cdot \bn_{v'}(\bq_{p,v} ))^2 \nonumber
\end{align}
\vskip -2.5mm
where $w_3$ and $w_4$ are weights both set to $0.3$.
We note that if a 3D point is projected onto a pixel that is masked as background (thus, its depth is invalid), then we exclude that pixel from the above summation. If the 3D\ point is projected onto background pixels in the majority of views, then this means that the point is likely an outlier and we remove it from the point cloud.
As a result, there are few $(p,p')$ pixel pairs in the above equation:\ each foreground pixel often has 3-4 corresponding pixels in other views.

\vspace{-1mm}
\paragraph{Energy minimization} The  energy is quadratic in the unknown pixel depths, thus we can minimize it by solving a linear system.
Due to the orthogonality term, which involves a linear combination (filtering) of depths within a pixel neighborhood, the depth of each pixel cannot be solved independently of the rest of the pixels. The solution can be computed through a sparse linear system - we provide its solution in our supplementary material. When we compute  the pixel depths, the corresponding 3D point positions, generated by these pixels, are updated. Given new 3D point positions, the consistency term also needs updating since the points might now be projected onto different pixels. This gives rise to an iterative  scheme, where at each step we estimate pixel depths by solving the linear system, then update the 3D point positions. We observed that the depth estimates become increasingly consistent across different views at each iteration and practically  convergence is achieved after 3-5 iterations. 
As shown in Figure \ref{fig:fusion}, the resulting point cloud  yields a  smoother reconstructed surface.

\vspace{-1mm}    
\paragraph{Mesh reconstruction and fine-tuning}
We apply the screened Poisson Surface Reconstruction algorithm \cite{Kazhdan:2013:SPS} to convert the resulting point cloud and normals to a surface mesh. Our method can optionally further ``fine-tune'' the generated mesh so that it matches the input contours more precisely. To do this, for each input line drawing we first extract its external contours and discretize them into a dense set of 2D\ points. Then for each input view, we render the mesh under the same orthographic projection, and find nearest corresponding mesh points to each contour point under this projection. Then we smoothly deform the 3D\ mesh such that the projected mesh points move towards the contour points under the constraint that the surface Laplacians \cite{Nealen:2005:SID}, capturing   underlying  surface details, are preserved. We also deform the mesh so that it better matches  the internal contours of the sketch. This is done by finding nearest corresponding mesh points to each internal contour point and scaling their Laplacian according to the scheme proposed in \cite{Nealen:2005:SID}. Mesh deformation is executed by solving a  sparse linear system involving all constraints from all internal and external contours across all input views. Figure \ref{fig:architecture} shows a reconstructed mesh before and after fine-tuning.
\vspace{-1mm}
\paragraph{Implementation} The network is implemented in Tensorflow \cite{tensorflow2015-whitepaper}. Training takes about $2$ days for 10K training meshes (40K training sketches) on a TitanX GPU. We use the Adam solver \cite{Adam} (hyperparameters $\beta1$ and $\beta2$ are set to $0.9$ and $0.999$ respectively). At test time, processing input sketches through the network takes 1.5 sec on a TitanX GPU, fusing the depth and normal maps takes 3 sec, mesh reconstruction and fine-tuning takes  4 sec (fusion and mesh reconstruction are implemented on the CPU - running times are reported on a dual Xeon E5-2699v3). In total, our method takes about 10 seconds to output  a shape.
Our source code and datasets are available on our project page:\\
{\footnotesize \url{https://people.cs.umass.edu/~zlun/SketchModeling}}


\vspace{-1mm}
\section{Evaluation}
\label{sec:evaluation}
\begin{table}[b!]
\centering
\small
\vspace{-5mm}
\begin{tabular}{c||c|c|c}
          & \#training  shapes & view A & view B \\
\hline
Character & 10000 & front & side \\
Airplane  & 3667  & top & side \\
Chair     & 9573  & front & side \\
\end{tabular}
\caption{Training dataset statistics.}
\label{table:dataset}
\end{table}

\begin{figure*}[t]
\centering
\includegraphics[width=1.0\linewidth]{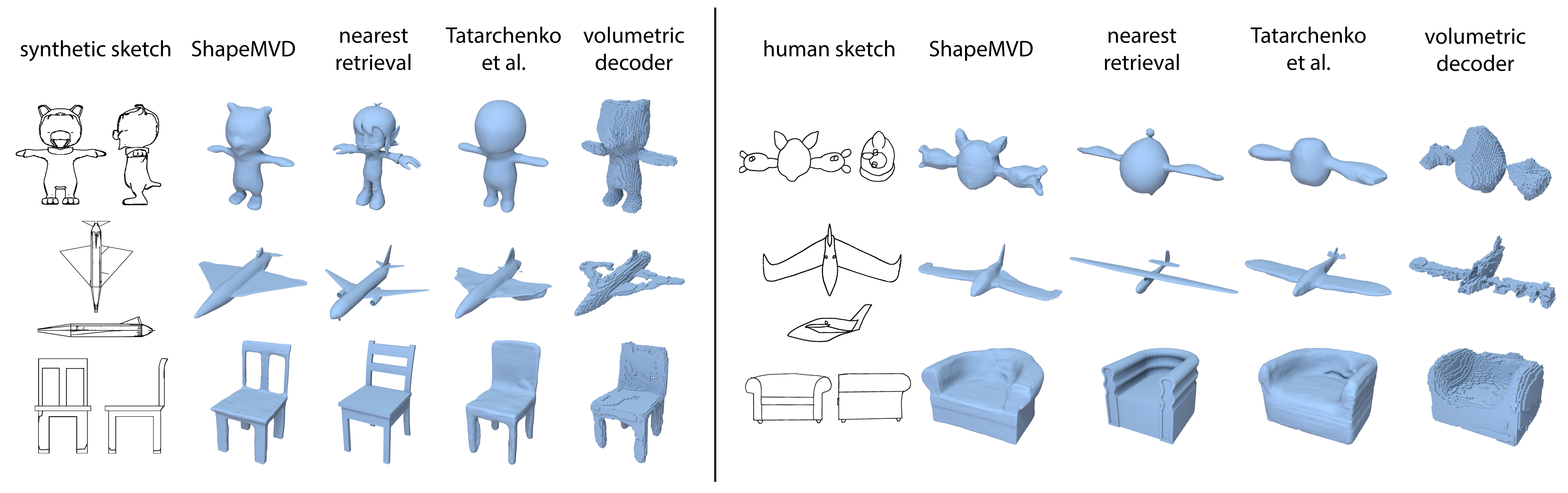}
\vspace{-10mm}
\caption{Comparisons of shape reconstructions from sketches for our method and baselines. }
\vspace{-1mm}
\label{fig:comparisons}
\end{figure*}

\begin{figure*}
\centering
\includegraphics[width=\linewidth]{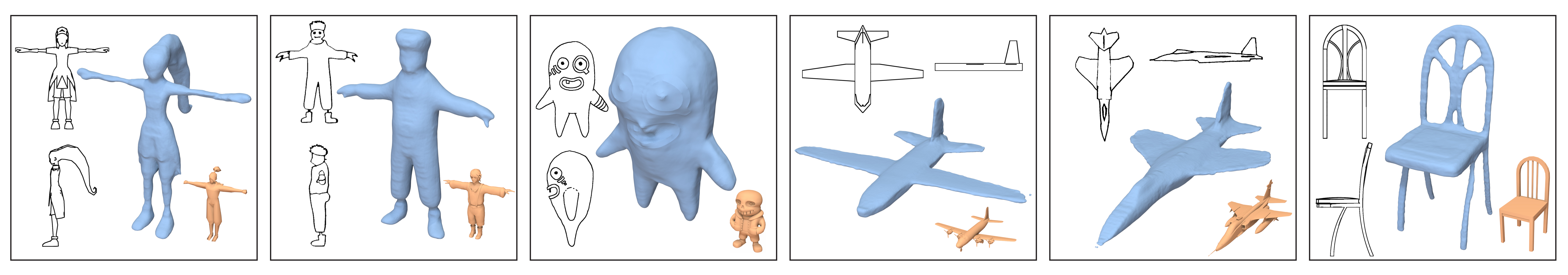}
\vspace{-8mm}
\caption{Gallery of results. Blue shapes represent reconstructions produced by our method from the  input sketches. Orange shapes are the nearest shapes in the training datasets retrieved via sketch-based retrieval. }
\vspace{-5mm}
\label{fig:gallery}
\end{figure*}
\vspace{-2mm}
We now discuss the experimental evaluation  of our method.
\vspace{-5mm}
\paragraph{Datasets} To train our network, we gathered three collections of 3D shapes along with their synthetic sketches. Each of the  collections included shapes belonging to the same broad category. The categories were 3D computer characters, airplanes, and chairs. To create the 3D computer character collection, we downloaded freely available 3D models of characters from an online repository (``The Models Resource'' \cite{modelsresource}). The collection contained humanoid, alien, and other fictional 3D\ models of characters. The airplanes and chairs originated from  3D ShapeNet \cite{shapenet}. 
We used these   particular categories from ShapeNet because the shapes in these categories have large geometric and structural variation. 
Table \ref{table:dataset} reports the number of training shapes and   view setting  used to generate the  training sketches.
\vspace{-1mm}
\paragraph{Test dataset} To evaluate our method and compare it with alternatives, we created a test dataset of synthetic and human line drawings for each of the above categories. Each  line drawing was created according to  a reference test shape. The goal of the evaluation was to examine how well the reconstructed 3D\ shapes  from these test line drawings matched the  reference test shapes.
To execute a proper evaluation, the reference test shapes should be sufficiently different from all training shapes. Otherwise, by overfitting a network to the training dataset or by simply using a nearest neighbor sketch-based retrieval approach, one could perfectly\   reproduce the reference shapes. To create the test dataset of reference shapes,
one option would be to randomly split the above collections
into a training and test part.
However, a problem with this strategy is that several test shapes would be overly similar to one or more training shapes because of  duplicate, or near-duplicate, 3D\ models that often exist in these collections (i.e., models that are identical up to an affine transformation, having tiny part differences or  different mesh resolution). To create our test dataset,
we found 120  shapes (40 per category) in our collections
that we ensured to be sufficiently different from the  shapes used for training by performing two  checks. First, for each shape, we  aligned it to each other shape in the collection through the best matching affine transformation and compute their Chamfer distance. The Chamfer distance
is computed by measuring the distance of each of the points on one shape  to the nearest surface point on the other shape, then the average of  these distances is used (we sampled 10K\ points uniformly per shape). We verified that the Chamfer distance between each test  shape and its nearest training shape is well above a threshold. Second, we rendered synthetic sketches for each  shape based on the input views  per category and extracted the representation from our encoder for these sketches. We then retrieved the   nearest other shape based on Euclidean distance over  the sketch representations. We verified that the distance is well above a threshold. We also visually confirmed that  test
and  training shapes
were different and the selected thresholds were appropriate.

For our 120 test shapes, we produced synthetic sketches for 90 of them (30 per category), and gathered human line drawings for the remaining 30 shapes (10 per category).  Synthetic sketches were produced from  the  test shapes using the line rendering techniques described in Section \ref{sec:training} based on the input views\ A and B  per category (Table \ref{table:dataset}). The human sketches were produced by asking two artists to provide us with hand-drawn line drawings of  reference test shapes.
The test shapes were presented to the artists on a computer display and were rendered using Phong shading. Their views were selected to approximately match the input views A and B per category. We asked the artists to create on paper line drawings  depicting the  presented shapes based on  the selected views.  We then scanned their line drawings, cropped and scaled them so that the scanned drawn area  matches the drawing area of training sketches on average.
In contrast to synthetic sketches, human line drawings tend to be noisy and inconsistent across different views.

\vspace{-1mm}
\paragraph{Evaluation measures} Given the above test  sketches as input, the goal of our evaluation is to measure how well the 3D\ shapes reconstructed by various methods, including ours, matched the  reference test shapes used to produce these sketches. Our method and the alternatives, listed in the following paragraphs, were trained and tested separately on each  category using the same splits.  We used five evaluation measures to compare the reconstructed shapes to the reference ones:\ Chamfer distance, Hausdorff distance, surface normal distance,  depth map error, volumetric Jaccard distance. 
The Hausdorff distance is computed by measuring the distance of each surface point on the reconstructed shape  to the nearest surface point on the reference shape, then computing the maximum of  these distances.
The surface normal distance 
is computed by measuring the angle between the  normal at each surface point on the reconstructed shape and the  normal  at the nearest surface point on the reference shape, then computing the  mean of the angles. 
The depth map error is computed by measuring the absolute differences between pixel depths in each of the output depth maps produced by our network and the corresponding depth maps of the reference shape, then  computing the average depth differences. To compute the volumetric Jaccard distance, we voxelized the reconstructed and reference shapes in a $128 \times 128 \times 128$ binary grid and measured the number of  voxels commonly filled in both shapes (their volume intersection) divided by the number of  their filled voxels (union of their volumes) - this is the Intersection over Union ($IoU$). We use $1-IoU$ as the volumetric Jaccard distance. 

\begin{table*}[h]
\centering
\small

\newcommand{\ShapeMVD}{{\!\!\!\footnotesize ShapeMVD}}
\begin{tabular}{@{}c@{}||c@{\,}|@{\,}c@{\,}|@{\,}c@{\,}|@{\,\,}c@{\,\,}|@{\,}c@{\,}|@{\,\,}c@{\,}||
                         c@{\,}|@{\,}c@{\,}|@{\,}c@{\,}|@{\,\,}c@{\,\,}|@{\,}c@{\,}|@{\,\,}c@{}}
     & \multicolumn{6}{c||}{Man-made objects (synthetic)} & \multicolumn{6}{c}{Character models (synthetic)} \\
     &           & nearest   & Tatarchenko                       & \cite{tatarchenko2016multi}+ & volumetric & R2N2
     &           & nearest   & Tatarchenko                       & \cite{tatarchenko2016multi}+ & volumetric & R2N2              \\
     & \ShapeMVD & retrieval & et al.\cite{tatarchenko2016multi} & U-net                        & decoder    & \cite{choy20163d}
     & \ShapeMVD & retrieval & et al.\cite{tatarchenko2016multi} & U-net                        & decoder    & \cite{choy20163d} \\
\hline
Hausdorff distance    & \textbf{0.092} & 0.165 & 0.142 & 0.121 & 0.113 & 0.144      & \textbf{0.089} & 0.200 & 0.119 & 0.092 & 0.152 & 0.148 \\
Chamfer distance      & \textbf{0.015} & 0.025 & 0.022 & 0.017 & 0.021 & 0.026      & \textbf{0.015} & 0.036 & 0.025 & 0.016 & 0.026 & 0.032 \\
normal distance       & \textbf{30.66} & 42.57 & 35.58 & 32.32 & 49.40 & 48.78      & \textbf{30.61} & 44.93 & 34.98 & 31.00 & 53.84 & 53.13 \\
depth map error       & \textbf{0.026} & 0.049 & 0.039 & 0.030 & 0.038 & 0.045      & \textbf{0.018} & 0.040 & 0.030 & 0.019 & 0.031 & 0.036 \\
volumetric distance\, & \textbf{0.344} & 0.501 & 0.442 & 0.374 & 0.432 & 0.512      & \textbf{0.313} & 0.541 & 0.428 & 0.329 & 0.437 & 0.493 \\
\end{tabular}

\begin{tabular}{@{}c@{}||c@{\,}|@{\,}c@{\,}|@{\,}c@{\,}|@{\,\,}c@{\,\,}|@{\,}c@{\,}|@{\,\,}c@{\,}||
                         c@{\,}|@{\,}c@{\,}|@{\,}c@{\,}|@{\,\,}c@{\,\,}|@{\,}c@{\,}|@{\,\,}c@{}}
     & \multicolumn{6}{c||}{Man-made objects (human drawing)} & \multicolumn{6}{c}{Character models (human drawing)} \\
     &           & nearest   & Tatarchenko                       & \cite{tatarchenko2016multi}+ & volumetric & R2N2
     &           & nearest   & Tatarchenko                       & \cite{tatarchenko2016multi}+ & volumetric & R2N2              \\
     & \ShapeMVD & retrieval & et al.\cite{tatarchenko2016multi} & U-net                        & decoder    & \cite{choy20163d}
     & \ShapeMVD & retrieval & et al.\cite{tatarchenko2016multi} & U-net                        & decoder    & \cite{choy20163d} \\
\hline
Hausdorff distance    & \textbf{0.116} & 0.176 & 0.153 & 0.153 & 0.130 & 0.149      & \textbf{0.117} & 0.188 & 0.139 & 0.136 & 0.178 & 0.168 \\
Chamfer distance      & \textbf{0.017} & 0.031 & 0.024 & 0.025 & 0.022 & 0.028      & \textbf{0.021} & 0.036 & 0.025 & 0.024 & 0.032 & 0.036 \\
normal distance       & \textbf{27.04} & 40.96 & 32.40 & 30.45 & 48.32 & 48.12      & \textbf{33.44} & 43.81 & 36.11 & 34.74 & 54.91 & 54.29 \\
depth map error       & \textbf{0.021} & 0.042 & 0.033 & 0.032 & 0.032 & 0.042      & \textbf{0.026} & 0.040 & 0.031 & 0.027 & 0.037 & 0.040 \\
volumetric distance\, & \textbf{0.311} & 0.544 & 0.405 & 0.403 & 0.405 & 0.500      & \textbf{0.298} & 0.458 & 0.342 & 0.307 & 0.420 & 0.436 \\
\end{tabular}
\caption{Comparisons of our method with baselines based on our evaluation measures (the lower the numbers, the better)}
\label{table:baseline}
\vspace{-6mm}
\end{table*}

\vspace{-2mm}
\paragraph{Comparisons} We  tested the reconstructions produced by our method (called ``ShapeMVD'') versus the following  methods: (a) a network based on the same encoder as ours but
using a volumetric decoder baseline instead of our multi-view decoder, (b) a network based on the same encoder as ours but
 with the
Tatarchenko \etal's view-based decoder \cite{tatarchenko2016multi} instead of our multi-view decoder, (c) the convolutional 3D LSTM architecture (R2N2) provided by Choy \etal~'s implementation \cite{choy20163d}, and (d) nearest sketch-based shape retrieval. For the volumetric decoder baseline (a), we used a $128 \times 128 \times 128$ output binary grid (the maximum we could fit in 12GB GPU\ memory). To make sure that the comparison is fair, we set the number of parameters in the volumetric decoder such that it is comparable to the number of parameters in our decoder. The volumetric decoder  consisted of five transpose 3D convolutions of stride $2$ and kernel size $4 \times 4 \times 4$. The number of filters  starts with $512$ and is divided by $2$ at each layer.  Leaky ReLU functions and batch normalization were used after each layer. We note that we did not use  skip-connections (U-net architecture) in the volumetric decoder because the size of the feature representations produced in the sketch image-based encoder is incompatible with the ones produced in the decoder. For Tatarchenko \etal's method,  the viewpoint is encoded into a continuous $64 \times 1$  representation  passed as input to the view-based decoder described in ~\cite{tatarchenko2016multi} without separate branches. To ensure a fair comparison, we increased the number of filters per  up-convolutional layer by a factor of $3$ so that the number of parameters in their and our decoder is comparable.   \rev{We also train it with the same loss function as ours. We additionally implemented a variant of Tatarchenko \etal's decoder by adding U-net connections between the encoder and their decoder. We report the evaluation measures on this additional variation.}
For the nearest-neighbor baseline, we  extract the representation of the input test sketches based on our encoder. This is used as a query representation to retrieve the training shape whose sketches have the nearest  encoder representation  based on Euclidean distance. All methods had access to the same training dataset per category and were evaluated on the same test set.

\vspace{-3mm}
Table  \ref{table:baseline} reports the evaluation measures for all competing methods based on both  synthetic and human  line drawings. We include evaluation separately for  organic shapes (3D\ character collection) and man-made shapes (measures  are averaged over airplanes and chairs). We also include standard deviations in our supplementary material. Our method produces much more accurate reconstructions than the competing methods in all cases. We note that  mesh fine-tuning was  not used here for any of the methods.
The reason was to evaluate the  methods by factoring out the post-processing effects of fine-tuning.
Fine-tuning is optional and does not significantly affect  the errors. It is used only to add details (``stylize'') the produced meshes based on the input contours when these are precisely drawn, and if users desire so (we provide more discussion regarding the effects of fine-tuning on the evaluation measures in the supplementary material). \rev{With respect to Tatarchenko \etal's method, we find that its enhancement with  U-net connections  improves its performance, but still performs worse than our  method, especially for man-made objects. This implies that  U-net  is a significant enhancement. We finally observe that the R2N2 does not perform better than our volumetric decoder baseline.}
Figure \ref{fig:comparisons} shows representative input test sketches, and output meshes for  competing methods (again, no fine-tuning is used here). 
In general, the nearest neighbor results  look plausible because retrieval returns human-modeled training shapes with fine details (e.g., facial features). Such details are not captured by any of the methods, including ours. 
On the other hand, as shown in the figure, and  confirmed by numerical evaluation, compared to nearest neighbor retrieval and other methods, ours produces shapes that better match the input sketch. The main reason is that our method better preserves the shape structure, topology and coarse geometry depicted in the input sketch.
From this aspect, we believe that the  shapes reconstructed by our method may serve as better starting ``proxies'' for artists to further improve upon. We also conducted an Amazon Mechanical Turk user study to perceptually evaluate the results of the methods. Specifically,
 we asked human subjects to compare the produced shapes from  different methods and select the one that best matches the input sketch. Our method was chosen by human participants to be the one producing shapes that best match the input sketches most of the time compared to the other methods including nearest retrieval (see supplementary material for results and more details on our user study).

\vspace{-1mm}
\paragraph{More results} The supplementary material includes all  reconstructed test shapes for our  method, nearest neighbors and competing methods (fine-tuning is not used on any of these results). We also include evaluation of our method against degraded variants  (e.g., using depth only, skipping the fusion step or GAN), and results using two sketches versus one sketch as input.  Figure \ref{fig:gallery} shows  shapes produced by our method for various input synthetic and human sketches. Fine-tuning was used for the meshes of this figure.


\vspace{-1mm}
\section{Conclusion}
\label{sec:conclusion}
\vspace{-2mm}
We presented an approach for 3D shape reconstruction from sketches.
Our method employs a ConvNet to predict depth and normals from a  set of viewpoints, and the resulting information is consolidated into a 3D point cloud via energy minimization. 
We evaluated our method and variants on two qualitatively different categories (characters and man-made objects). Our results indicate that view-based reconstruction of a 3D shape is significantly more accurate than voxel-based reconstruction. We also showed that our method can generalize  to human-drawn sketches. We believe that there is significant room for  improving our method in the future. For example, it would be interesting to explore the possibility of incorporating the fusion process  in the network, and modifying its architecture  such that reconstruction is done from  arbitrary viewpoints. Our  reconstructed shapes often lack fine details that users would prefer to see in production-quality 3D\ models.
We believe that these shapes  can serve as  starting ``proxies'' for artists to  improve upon through modeling interfaces. From this aspect, it would be useful to integrate interactive modeling techniques into our method.

\vspace{-2mm}
\section*{Acknowledgements}
\vspace{-2mm}
We acknowledge support from NSF (CHS-1422441,CHS-1617333,IIS-1617917,IIS-1423082), Adobe, NVidia, Facebook. We  acknowledge the MassTech Collaborative grant for funding the UMass GPU cluster.


{
\bibliographystyle{IEEEtran}
\bibliography{sketch_modeling}
}

\end{document}